\documentclass[runningheads]{llncs}

\usepackage{eccv}

\usepackage{eccvabbrv}

\usepackage{graphicx}
\usepackage{booktabs}

\usepackage[accsupp]{axessibility}  

\usepackage{hyperref}

\usepackage{orcidlink}

\usepackage{booktabs}
\usepackage{xcolor}
\usepackage{pifont}
\usepackage{array} 
\newcommand{\cmark}{\textcolor{green!60!black}{\ding{51}}}
\newcommand{\xmark}{\textcolor{red!80!black}{\ding{55}}}
\usepackage[table]{xcolor} 
\definecolor{headerbg}{gray}{0.9} 
\usepackage{multirow}
\usepackage{colortbl}
\usepackage{pifont}
\usepackage{makecell} 
\usepackage{xcolor}
\usepackage{booktabs}
\usepackage[utf8]{inputenc}
\usepackage{multirow}
\usepackage{graphicx}
\usepackage[table]{xcolor}
\usepackage{tabularx} 
\usepackage{bm}
\usepackage{wrapfig}
\usepackage{makecell}
\usepackage{multirow}

\begin{document}

\title{OmniHuman: A Large-scale Dataset and Benchmark for Human-Centric Video Generation} 

\titlerunning{OmniHuman}




\author{
Lei Zhu$^{*}$\inst{1} \and
Xing Cai$^{*}$\inst{2} \and
Yingjie Chen\inst{2} \and
Yiheng Li\inst{3} \and
Binxin Yang\inst{2} \and
Hao Liu\inst{2} \and
Jie Chen\inst{1} \and
Chen Li\inst{2} \and
Jing LYu\inst{2}
}

\authorrunning{L.~Zhu et al.}

\institute{
Peking University, Beijing, China \and
WeChat Lab, Tencent, China \and
Chinese Academy of Sciences, Beijing, China\\
$^{*}$Equal contribution.
}

\maketitle

\begin{abstract}
Recent advancements in audio-video joint generation models have demonstrated impressive capabilities in content creation. However, generating high-fidelity human-centric videos in complex, real-world physical scenes remains a significant challenge. We identify that the root cause lies in the structural deficiencies of existing datasets across three dimensions: limited global scene and camera diversity, sparse interaction modeling (both person-person and person-object), and insufficient individual attribute alignment.
To bridge these gaps, we present \textbf{OmniHuman}, a large-scale, multi-scene dataset designed for fine-grained human modeling. OmniHuman provides a hierarchical annotation covering video-level scenes, frame-level interactions, and individual-level attributes. To facilitate this, we develop a fully automated pipeline for high-quality data collection and multi-modal annotation. Complementary to the dataset, we establish the OmniHuman Benchmark (\textbf{OHBench}), a three-level evaluation system that provides a scientific diagnosis for human-centric audio-video synthesis. Crucially, OHBench introduces metrics that are highly consistent with human perception, filling the gaps in existing benchmarks by providing a comprehensive diagnosis across global scenes, relational interactions, and individual attributes. 
Experiments show that fine-tuning on only 20\% of OmniHuman significantly boosts performance, validating its effectiveness in advancing complex scenario modeling. The code is available at \url{https://github.com/julia-cherry/OmniHuman}.

\keywords{Audio-visual dataset \and Human-centric video generation}

\end{abstract}
\section{Introduction}
\label{sec:introduction}

\begin{figure}[h]
    \vspace{-0.3cm}
    \centering
    \includegraphics[width=1.\linewidth]{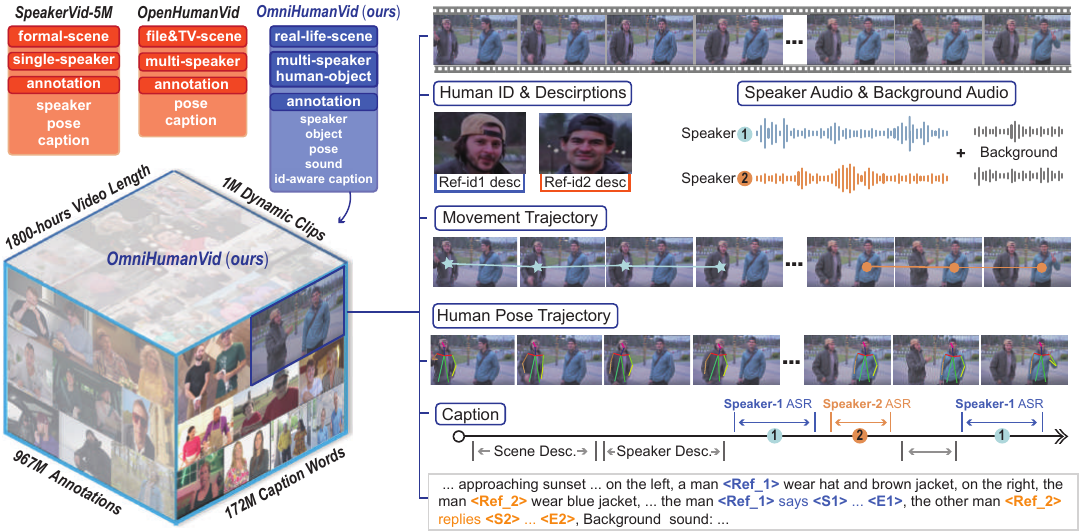}
    \vspace{-1.8em}
    \captionof{figure}{
    \textbf{OmniHuman}: a 1M-video, 1800 hours, 80K-identity dataset with hierarchical annotations covering diverse natural scenes and social interactions.
    }
    \label{fig:motivation}
    \vspace{-0.8cm}
\end{figure}

The new generation of video-audio joint generation models, represented by Sora 2\cite{openai2025sora2}, ovi\cite{ovi} and Veo 3.1\cite{veo3.1}, are gaining significant attention and are widely applied in downstream tasks such as film production, social media, and interactive entertainment. Human behavior, serving as the visual focus and narrative core, carries extremely high semantic density and presents unprecedented challenges for the fine-grained control of generative models. Although existing models have achieved promising results in generating single-person close-ups or isolated actions, their ability to generalize to complex, real-world physical scenes remains severely hindered. We argue that the root cause lies in the systematic and structural deficiencies of current human-centric datasets across core dimensions. Specifically, these deficiencies are manifested in the following three levels.


\textbf{\textit{(i) Global: Scene and Camera Diversity.}} Existing human-centric video datasets are mostly limited to controlled environments (e.g., formal or semi-formal scenes), causing poor generalization to complex real-world scenes. Most also lack ambient audio and environment annotations, hindering models from learning the coupling between human actions and environmental sounds, leading to semantic audio-visual mismatches in generation. Limited diversity in shot scales and camera movements further induces artifacts like feature instability and background distortion in specific shot types.
\textbf{\textit{(ii) Interaction: Person-Person and Person-Object.}} Current datasets focus predominantly on single-person videos, with multi-person interactions accounting for less than 3\% \cite{intergen}. This imbalance results in three key deficiencies in multi-person scenarios: relational distortion (lack of authentic social dynamics), audio-visual misassignment (failure to map speech to the correct speaker), and identity drift (facial feature blending or misalignment). Moreover, videos with person-object interactions such as precise hand-tool contact—are scarce, limiting the modeling of physical interactions and narrowing the range of generated behaviors.


To address these challenges, we built an automatic pipeline to collect and annotate high-quality video and audio data. The annotations cover three levels: video-level scene information, frame-level interactions, and individual-level attributes. This led to the creation of \textbf{OmniHuman}, a multi-scene dataset for detailed human modeling. A comparison with other datasets is shown in Table~\ref{tab:dataset_comparison}. Our dataset introduces new features at three levels:
\textbf{\textit{(i)Global Level}}: OmniHuman includes most common real-life scenes. We provide clear labels for scene types and background sounds. This helps models learn to generate videos where people and their surroundings match in both look and sound. The videos also include different camera angles and both static and moving shots.
\textbf{\textit{(ii)Interaction Level}}: We built a large collection of videos showing two people interacting—talking, working together, or competing. We label each person's actions and identity, so models can learn who does what in group scenes, avoiding issues like mixed-up identities or unnatural interactions. The dataset also includes videos where people interact with objects, expanding from social to physical interactions.
\textbf{\textit{(iii)Individual Level}}: We selected high-quality videos with good lip-sync and clear visuals. For each person, we provide detailed labels: video-side labels include person ID, body movements, and face/hand clarity; audio-side labels include speech text, speaker ID, emotion, and lip-sync quality. 

Based on the above hierarchical analysis and dataset construction, we establish a corresponding benchmark—OmniHuman Benchmark (\textbf{OHBench}). Existing benchmarks \cite{vabench, hua2025vabench} have two key limitations: metrics misaligned with human perception and missing critical dimensions. At the scene level, they lack evaluation of background plausibility across shot types and semantic audio-visual consistency. At the interaction level, they miss assessments of interaction naturalness, audio-visual assignment accuracy, and human-object interaction plausibility. At the individual level, subject consistency and speech quality are overlooked. These gaps hinder accurate model evaluation in complex real-world scenarios. To address this, we select diverse samples from our dataset with domain gaps relative to the training data and construct a three-level evaluation system for comprehensive model diagnosis.
Through this hierarchical construction—from global scenes to interaction relationships and then to individual attributes—the OmniHuman dataset and its corresponding benchmark form a closed loop. Specifically, the main contributions of this work are threefold:
\begin{itemize}
\setlength{\itemsep}{0pt} 
    \item We construct \textbf{OmniHuman}, a large-scale, multi-scene audio-visual dataset. Its effectiveness is underscored by substantial performance gains achieved by fine-tuning an open-source model on a mere 20\% of the data.
    \item We develop a fully automated hierarchical fine-grained annotation pipeline that caters to the diverse needs of downstream tasks across video, frame and individual levels.
    \item We establish \textbf{OHBench}, a comprehensive benchmark aligned with human perception for human-centric  multimodal video generation, enabling comprehensive diagnosis of model performance in real-world scenarios.
\end{itemize}

\section{Related Work}
\label{sec:related}

\subsection{Human-Centric Video Generation}

\begin{table}[t]
\centering
\caption{Comparison of different mainstream datasets. Columns are color-coded by levels: Global, Interactional, and Individual.}
\vspace{-1.0em}
\label{tab:dataset_comparison}
\resizebox{\columnwidth}{!}{%
    \setlength{\tabcolsep}{3pt} 
    \renewcommand{\arraystretch}{1.2} 
    
    \begin{tabular}{l | cccc | cc | ccc}
    \toprule
    \textbf{Dataset} & 
    \cellcolor{blue!15}\textbf{Multi-scene} & \cellcolor{blue!15}\textbf{Shot-type} & \cellcolor{blue!15}\textbf{Camera} & \cellcolor{blue!15}\textbf{Sound} & 
    \cellcolor{green!15}\textbf{person-person} & \cellcolor{green!15}\textbf{person-object} & 
    \cellcolor{orange!15}\textbf{Speaker-anno} & \cellcolor{orange!15}\textbf{Pose-anno} & \cellcolor{orange!15}\textbf{Object-anno} \\
    \midrule
    ActivityNet\cite{caba2015activitynet}   & \cmark & \cmark & \xmark & \xmark & \xmark & \xmark & \xmark & \xmark & \xmark \\
    TikTok-v4 \cite{di2023magicdance}    & \cmark & \cmark & \xmark & \xmark & \xmark & \xmark & \xmark & \cmark & \xmark \\
    Openhumanvid \cite{li2025openhumanvid}  & \cmark & \cmark & \cmark & \xmark & \cmark & \xmark & \xmark & \cmark & \xmark \\
    \midrule
    CelebV-HQ \cite{celebV}    & \xmark & \xmark & \xmark & \cmark & \cmark & \xmark & \xmark & \xmark & \xmark \\
    VoxCeleb2 \cite{voxceleb2}   & \cmark & \xmark & \xmark & \xmark & \xmark & \xmark & \xmark & \xmark & \xmark \\
    HDTF  \cite{HDTF}   & \xmark & \xmark & \xmark & \xmark & \xmark & \xmark & \xmark & \xmark & \xmark \\ 
    SpeakerVid-5M  \cite{speakervid} & \xmark & \cmark & \cmark & \xmark & \xmark & \xmark & \cmark & \cmark & \xmark \\ 
    \midrule
    \textbf{OmniHuman (ours)} & \cmark & \cmark & \cmark & \cmark & \cmark & \cmark & \cmark & \cmark & \cmark \\ 
    \bottomrule
    \end{tabular}
}
\vspace{-0.5cm}
\end{table}

With the rapid development of deep learning, human-centric video generation has become increasingly important in practical applications\cite{peng2023selftalk, peng2025omnisync, peng2025dualtalk}.
Audio-visual synchronized video generation has evolved from cascaded systems to a native bimodal paradigm \cite{wang2024av,javisdit,WeaklyTVQA, peng2024synctalk}, jointly denoising visual and audio streams in a unified architecture. Early work like MM-LDM \cite{MM-LDM} explored multimodal latent diffusion for joint synthesis, while JavisDiT \cite{javisdit} enhanced alignment via hierarchical spatio-temporal prior estimation, though it struggles in complex human-centric scenes. Meanwhile, proprietary models such as Sora2, Veo3.1, and Wan2.5 \cite{wan2.5} set industry benchmarks in physical simulation and high-resolution one-pass generation. In the open-source human-centric domain \cite{apollo, uniavgen, jova}, Ovi uses symmetric twin-tower fusion, MOVA \cite{MOVA} employs mixture-of-experts, and LTX-2 \cite{LTX-2} adopts asymmetric dual-stream for efficiency and precision. For speech-to-video (S2V), recent breakthroughs include OmniAvatar \cite{gan2025omniavatar} with multi-hierarchical audio embeddings, MultiTalk \cite{sung2024multitalk} using Label Rotary Position Embedding for multi-character audio binding, and InfiniteTalk \cite{yang2025infinitetalk} enabling infinite-length generation via sparse-frame dubbing.

\subsection{Evolution of Human-Centric Datasets and BenchMarsks}

Human-centric video datasets and evaluation benchmarks have evolved in synergy, transitioning from early constrained portraits (e.g., VoxCeleb\cite{voxceleb2}, HDTF\cite{HDTF}) toward interactive scenarios (e.g., OpenHumanVid\cite{li2025openhumanvid}, SpeakerVid-5M\cite{speakervid}). Despite this progress, systematic gaps persist in scene diversity, holistic environment-aware acoustics, complex person-object physical interactions, and fine-grained attribute persistence. Correspondingly, while assessment paradigms have shifted from visual aesthetics (VBench\cite{vbench}) toward multi-dimensional semantic and physical diagnostics (VABench\cite{hua2025vabench}, SVBench\cite{yang2025svbench}, MTAVG-Bench\cite{zhou2026mtavg}, PhyAVBench\cite{xie2025phyavbench}), current benchmarks remain insufficient for evaluating dyadic audio-visual assignment, background plausibility, and precise physical contact, often exhibiting discrepancies with human perception. To bridge these dual gaps, we propose OmniHuman, a hierarchically annotated dataset, and its companion OHBench, a three-tier evaluation system that enables a comprehensive diagnosis of model deficiencies in complex dynamic scenarios through strictly aligned, fine-grained evaluation dimensions.
\section{OmniHuman}
\label{sec:data curation}
\begin{figure}[t]
    \centering
    \includegraphics[width=1.\linewidth]{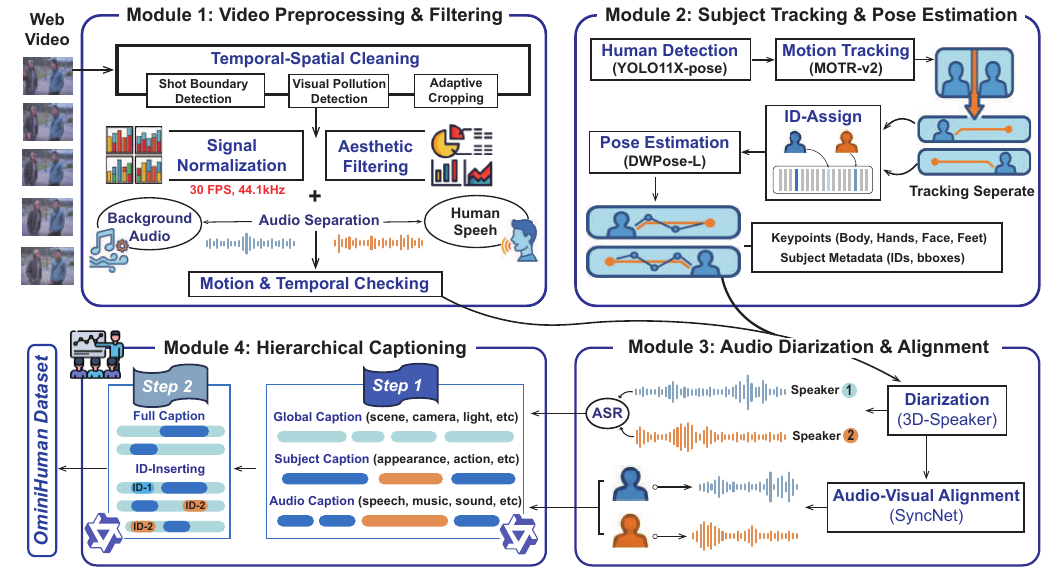}
    \vspace{-1.5em}
    \captionof{figure}{
    OmniHuman employs a fully automated pipeline for high-quality data collection and fine-grained annotation, with each module applying progressive filtering to ensure both video quality and annotation accuracy.
    }
    \label{fig:pipeline}
    \vspace{-1.5em}
\end{figure}

The construction process of OmniHuman consists of several decoupled modules: video preprocessing \& filtering, subject tracking and pose estimation, audio-visual alignment and synchronization verification, and hierarchical multi-modal caption generation.
Each module applies progressive filtering to ensure both video quality and annotation accuracy.
Through modular deployment, the full data pipeline operates at scale in a streaming manner.
The complete construction pipeline is illustrated in Fig.~\ref{fig:pipeline}.

\subsection{Video Preprocessing \& Filtering.}
\label{subsec:preprocessing}
This module forms the foundation of the data pipeline and aims to transform raw videos from open, complex scenarios into spatiotemporally consistent, high-availability segments with clean signals.
This module is decomposed into four hierarchical stages for progressive purification of multi-source video data.\\
\textbf{Temporal–spatial cleaning.}
Temporally, we obtain semantically coherent short clips through shot segmentation using the TransNetV2 \cite{Transnetv2} algorithm.
Spatially, we use OCR and logo detection algorithms to estimate text-contaminated regions, then crop frames to retain the main subject area, thereby removing visual pollutants such as subtitles and logos.
This process establishes foundational visual purification, improves frame quality, and ensures controllable clip duration. \\
\textbf{Audio governance and signal standardization.}
This layer ensures audio usability and consistency by removing samples with missing tracks, abnormal duration, or low quality (high silence ratio and low volume).
All clips are standardized to 30 FPS and 44.1 kHz to eliminate potential source clock offsets.
Demucs~\cite{Demucs} is then applied for four-source separation, extracting vocals as the target track and mixing the remaining tracks as background.
This design provides clean audio for subsequent audio-content processing tasks, including ASR and lip-audio alignment.\\
\textbf{Aesthetic filtering.}
This layer adopts a lightweight-to-heavy filtering strategy.
First, frame-level CLIP~\cite{clip,FreestyleRet} aesthetic scoring rapidly removes low-quality clips.
Next, OCR-based text-density analysis and watermark detection estimate the density and confidence of residual text, subtitles, and watermarks, and filter clips accordingly.
Furthermore, DOVER~\cite{DOVER} evaluates composition and clarity at the video level and filters clips based on the resulting scores.
This process further enhances visual cleanliness and video quality.\\
\textbf{Motion filtering and temporal consistency.}
Finally, we perform motion and temporal consistency checks.
First, we use UniMatch~\cite{yang2025unimatch} flow to analyze motion quality, retaining valid motion while filtering static, shaky, or abnormal segments.
Then, we enforce temporal continuity by combining perceptual hashing with SigLIP~\cite{sigclip} cosine similarity to filter semantically inconsistent frames.
The four-tier pipeline progressively optimizes spatial-temporal usability, audio purity, visual fidelity, and temporal semantic coherence.

\subsection{High-Fidelity Subject Tracking and Pose Estimation.}
\label{sec:tracking}
\textbf{Spatio-temporal Multi-Object Tracking.}
To capture human motion in interactive scenarios, we build a highly robust multi-object tracking pipeline.
We use YOLOv11~\cite{YOLOv11} to detect human instances, and feed NMS-refined detections into MOTRv2~\cite{MOTRv2} to model cross-frame associations via query propagation.
We allow a maximum loss span of 5 frames to handle short-term occlusion.
The pipeline outputs per-identity metadata as $B_i = \{(box_m, score_m)\}_{m=t_{start}}^{t_{end}}$, where $box_m$ and $score_m$ denote the detection box and confidence score at frame $m$.\\
\textbf{Fine-grained Whole-body Pose Representation.} 
Based on identity-coherent tracking, we perform fine-grained 2D whole-body pose estimation for each instance using DWPose-L for body, face, and feet, with dedicated hand detection and optimization.
The pipeline performs frame-wise detection of 134 full-skeletal keypoints for each tracked instance, denoted as \(\mathcal{K} = \{k_j\}_{j=0}^{133}\), where each keypoint \(k_j = (x_j, y_j, c_j)\) consists of 2D coordinates and confidence \(c_j \in [0, 1]\).
These keypoints cover body core, foot support, facial features, and both hands.
To ensure facial fidelity, we compute a clarity score \(C_s = \text{Var}(\Delta R)\) for face region \(R\) based on Laplacian variance, and remove videos with persistently sub-threshold facial clarity.
For identity assignment, we extract face embeddings(ArcFace~\cite{ArcFace}) from the frame with the highest average confidence of face keypoints in each track.
We assign the same ID to frames with embedding similarity above 0.55, yielding a unique person ID per video. 

\subsection{Audio-Visual Alignment and Synchronization Verification.} 
In human-centric scenarios, precise synchronization between audio and visual subjects is essential for a high-quality dataset.
This module first uses the vocal signal $A_{vocals}$ extracted in Section~\ref{subsec:preprocessing} and performs speaker diarization with 3Dspeaker~\cite{3Dspeaker}.
It identifies $M$ active speech intervals $\{[t_{start}, t_{end}]_m\}_{m=1}^M$ with unique speaker indices.
To match speaking intervals with visual identity trajectories while suppressing voice-over interference, we further introduce SyncNet\cite{syncnet} to resolve audio-visual attribution.
SyncNet takes face bounding-box sequences from Section~\ref{sec:tracking} as visual input and computes cross-correlation between face regions and each speech interval in a joint embedding space, producing a synchronization score $S_{sync}$.
The system then applies greedy matching to assign each audio segment to the visual ID trajectory with the highest response.
A sample is retained only when all detected subjects satisfy $S_{sync}$ above a preset threshold and the temporal offset is within 3 frames.
Unmatched audio segments are preserved as background audio.
We then apply ASR (FunASR-Nano~\cite{an2025funasrtechnicalreport}) to each retained audio segment to obtain detailed speech transcripts.
For each subject $i$, the matched synchronization metadata is defined as $\mathcal{S}_i = \{[t_{start}, t_{end}]_m\}_{m \in \mathcal{M}_i}$, where $\mathcal{M}_i$ denotes the segment subset associated with identity $i$.
The pipeline outputs a structured subject set $\mathcal{P} = \{(\text{ID}_i, B_i, \mathcal{K}_i, \mathcal{S}_i)\}_{i=1}^N$ as comprehensive spatiotemporal constraints.
This high-precision alignment provides a reliable physical basis for training audio-visual joint generation models and supplies clean cross-modal supervision for downstream speech-to-video generation tasks.



\subsection{Hierarchical Multi-modal Caption Generation}
\textbf{Hierarchical Feature Deconstruction and Semantic Fusion.} 
We employ Qwen3-Omni~\cite{qwen-omni} as the inference core to implement a two-stage generation strategy that covers global audio-visual context and detailed subject-level attributes.
In stage one, the MLLM captures global video context—video type, shot scale, camera motion, background, lighting and extracts fine-grained attribute sets $\mathcal{X}_i$ (appearance, motion, expression) for each subject $\text{ID}_i$.
Audio analysis covers speech, music, and sound effects: \textbf{1)} speech transcription $w_m$ with associated subject ID, emotion, and timestamps, each speech metadata is represented as $\mathcal{A}_i = \{([t_{start}, t_{end}]_m, w_m, \text{ID}_m, \text{emotion}_m)\}_{m \in \mathcal{M}_i}$; \textbf{2)} music attributes (type, mood, relative volume); \textbf{3)} sound effect categories with textual descriptions.
We enforce strict mutual exclusivity between speech and music to avoid overlapping supervision signals and maintain disentangled audio representations.
Additionally, we apply rigorous post-processing to deduplicate repeated speech and lyric content.

\begin{figure}[t]
    \centering
    \includegraphics[width=1.\linewidth]{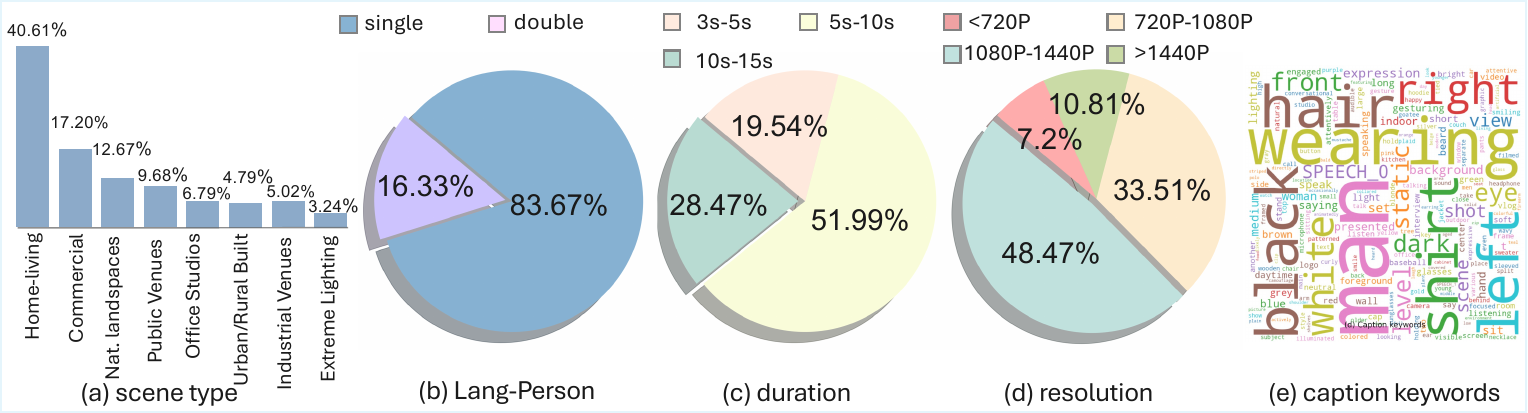}
    \vspace{-1.5em}
    \captionof{figure}{Statistical analysis of the OmniHuman dataset composition.
    }
    \label{fig:data_static}
    \vspace{-1.5em}
\end{figure}
In stage two, the model synthesizes fragmented attributes into a coherent long-form narrative caption $\mathcal{C}_i$.
To reduce hallucination, we introduce a placeholder mechanism.
Instead of generating exact dialogue or lyrics directly, the model inserts fixed-position anchors such as $[speech\_m]$ and $[lyrics\_m]$.
These anchors are later replaced with the corresponding transcribed content from stage one, preserving narrative fluency while ensuring strict consistency with the source video on critical information such as dialogue and lyrics.
To mitigate MLLM hallucinations and labeling errors, we introduce multiple validation checks. The number of structured subject labels must match the tracking module's output (Sec.~\ref{sec:tracking}), and the speaker count and transcribed content must remain within an acceptable edit-distance margin from upstream ASR transcripts. A video is retained only if it passes all checks. \\
\textbf{Identity-Aware Multi-subject Alignment and Insertion.}
To resolve identity ambiguity in multi-person scenes, we introduce an identity alignment mechanism based on reference face images.
Based on whole-body tracking results in Section~\ref{sec:tracking}, we select, for each subject, the frame with the highest average confidence over facial keypoints as the reference face.
During stage one, the system processes the video together with fixed reference faces $\{\text{REF}_1, \text{REF}_2\}$ and enforces explicit identity verification.
Each subject $i$ is assigned an identity label $\text{ID}_i \in \{\text{REF}_1, \text{REF}_2\}$, establishing robust feature-level correspondence.
In stage two semantic fusion, we adopt a two-step referential insertion strategy.
The system first generates the raw caption $\mathcal{C}_i$.
It then inserts identity anchors $\langle\text{REF}_i\rangle$ via suffix tagging after each noun or pronoun referring to that subject.
Ultimately, the pipeline outputs a structured subject set $\mathcal{P} = \{(\text{ID}_i, B_i, \mathcal{K}_i, \mathcal{X}_i, \mathcal{A}_i, \mathcal{C}_i)\}_{i=1}^N$ as comprehensive spatiotemporal and semantic constraints.

\subsection{OmniHuman Statistics and Analysis}
\label{sec:data curation}


As shown in Fig.~\ref{fig:data_static}, our dataset comprises 1 million videos totaling 1,800 hours, encompassing 80,000 distinct identities. 
The primary language is English, accounting for up to 80\%. 
It covers eight typical shooting scenarios: home living spaces, business offices and creative studios, natural environments and landscapes, urban and rural built landscapes, public performance and sports venues, commercial and retail spaces, and industrial and work areas, with detailed distributions illustrated in Fig.~\ref{fig:data_static}(a). 
Additionally, the dataset includes eight video content types: talk shows and interviews, educational, food and cooking, sports and fitness, music and performance, gaming, product reviews, and movies and TV. 
Based on the number of visual subjects, our dataset is further categorized into two classes, single-person and two-person, with their proportions shown in Fig.~\ref{fig:data_static}(b). 
The pie charts in Fig.~\ref{fig:data_static}(c) and (d) illustrate the distributions of video resolution and duration, respectively. 
Technically, all videos are of high-definition quality or higher, providing a robust visual foundation for model training. 

\section{OmniHuman Benchmark (OHBench)}
\label{sec:benchmark}

\begin{figure}[t]
    \centering
    \includegraphics[width=0.9\linewidth]{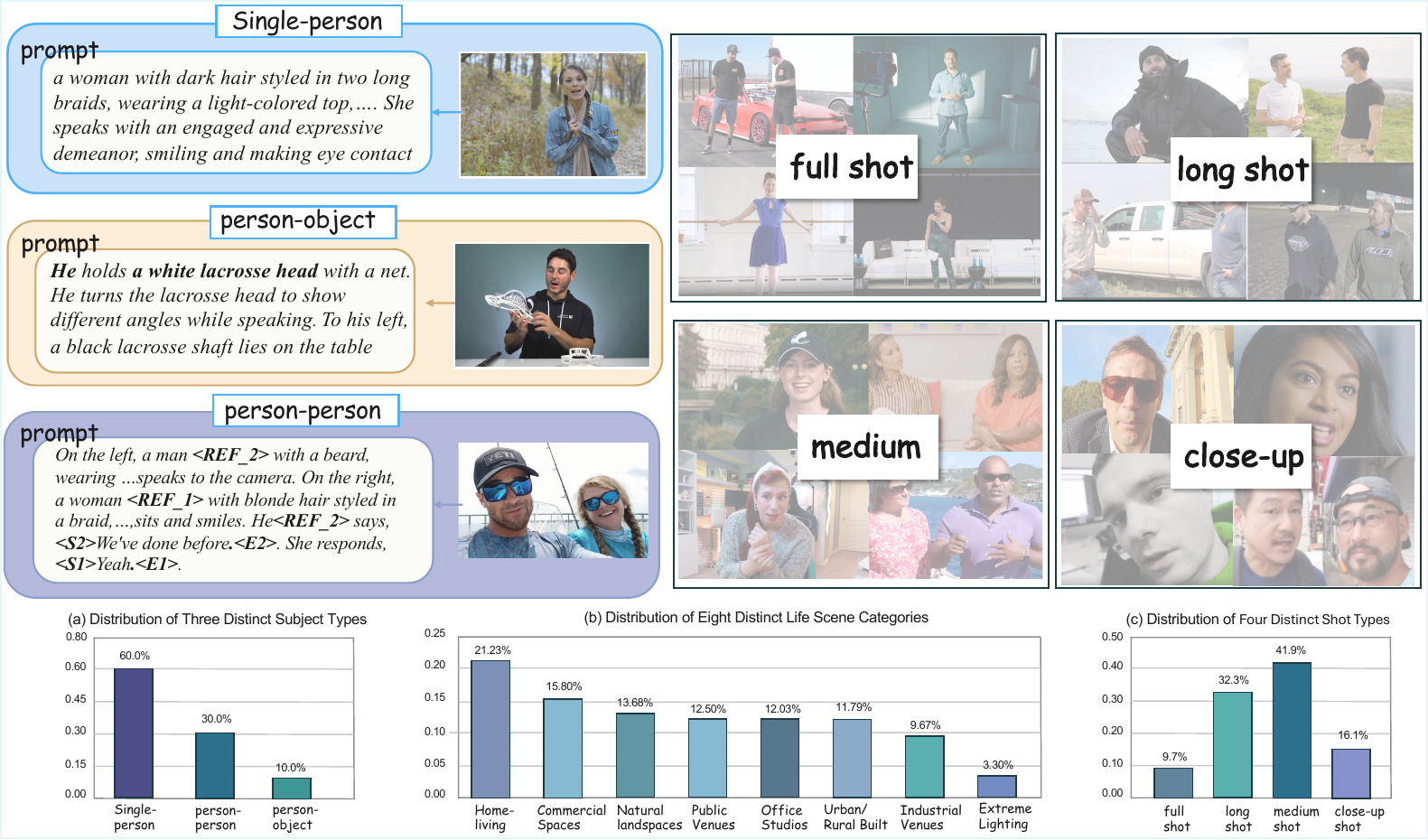}
    \captionof{figure}{Distribution of subject categories, scene types, and shot types in OHBench.
    }
    \label{fig:bench_static}
    \vspace{-0.5cm}
\end{figure}

\textbf{Data Collection of OHBench.}
Our evaluation set samples from OmniHuman with domain gaps relative to the training set; its distribution is shown in Fig.~\ref{fig:bench_static}. At the global level, we cover eight real-life scenarios across various shot scales (extreme long to extreme close-up). At the relational level, we include single and two-person videos at a fixed ratio, with a particular focus on challenging person-object interaction scenarios including tool use and physical manipulation. In total, the benchmark contains 331 single-person, 128 two-person, and 50 person-object interaction videos.\\
\textbf{Supported Tasks.}
The OmniHuman Benchmark offers broad task applicability, fully supporting various human-centric video generation tasks, including audio-video joint generation, speech to video, video dubbing, controllable human video editing and downstream speech generation tasks.  \\
\textbf{Overview of Metrics.}
OHBench is organized into three progressive levels with seven dimensions. At the global level, we assess video quality, audio quality, and multi-modal alignment. At the relational level, we evaluate person-person and person-object interaction. At the individual level, we measure subject-video and subject-audio fidelity. These dimensions enable comprehensive diagnosis of fine-grained human-centric scene modeling.
We find existing metrics (e.g., VBench's aesthetic score, identity consistency, and SyncFormer \cite{iashin2024synchformer} for audio-visual synchronization) often misalign with human perception. To address this, we construct a new evaluation suite by curating perception-aligned existing metrics and designing novel ones for previously uncovered dimensions. We provide evidence in the supplementary material demonstrating that our metrics better align with human judgment.
Our evaluation integrates expert model-based and MLLM-based assessments; all prompts are provided in the supplementary material.

\subsection{Foundational Synthesis Quality}


\textbf{Video Quality.} 
We evaluate video quality through two complementary aspects: foundational quality and background plausibility. For foundational quality, we adopt metrics from VBench aligned with human perception, focusing on imaging quality (IQ) and dynamic degree (DD). Specifically, we use MUSIQ \cite{MUSIQ} to predict per-frame imaging quality, capturing distortions like overexposure, noise, and blur, and RAFT \cite{RAFT} optical flow to assess motion intensity. For background plausibility, we leverage Gemini-3 to perform semantic analysis and physical plausibility judgment on video frames, assessing temporal realism—particularly temporal stability in static shots and parallax plausibility under dynamic camera movements. The model outputs a score from 1 to 10.\\
\textbf{Audio Foundation Quality.} We compute the aesthetic score (AbS) as the average of four Audiobox\cite{audiobox} dimensions: content enjoyment, content usefulness, production complexity, and production quality. Additionally, we introduce KL divergence and Fréchet Distance (FD) to measure the distributional and perceptual similarity between generated and real audio, respectively. \\
\textbf{Audio-Visual Synchronization and Semantic Consistency.} 
To evaluate audio-visual synchronization (A-V) in complex environments, we use ImageBind \cite{ImageBind} to measure temporal offsets between visual and auditory events, assessing synchronization accuracy between action sounds and background. Additionally, we employ CLAP \cite{CLAP} to compute audio-text semantic similarity (T-A), quantifying alignment between generated sound effects and corresponding text.

\subsection{Social-Physical Interaction Realism}
\textbf{Person-Person Interaction Quality.} 
For two-person interaction scenarios, we leverage the multi-modal large model Gemini-3-pro to evaluate interaction capabilities across three dimensions, each rated on a scale of 1 to 10:: interaction naturalness (IN) assesses social plausibility via eye contact, body gestures, and action responses (e.g., nodding, gaze following); listener realism (LR) evaluates both the accuracy of mapping audio signals to the corresponding speaker (multi-agent audio-visual assignment) and whether the silent listener maintains a natural expression without producing lip movements.; emotion similarity (ES) measures whether the emotional expressions of interacting parties are coordinated and align with semantic context and social expectations.\\
\textbf{Dual-person ID Consistency}. 
We employ ArcFace \cite{ArcFace} to extract identity embeddings for both individuals and compute the cosine similarity between these embeddings and their ground-truth counterparts. The average similarity score (IC$^*$) quantifies identity drift in dual-person scenarios. \\
\textbf{Person-Object Interaction Quality.} 
For person-object interaction, we evaluate two core dimensions using Gemini-3 (1–10, higher is better): object consistency (OC) measures the temporal stability of an object's appearance, position, and state, ensuring it remains identifiable and follows physical laws; contact naturalness (CN) assesses the physical plausibility of the human-object interface via spatial accuracy, temporal synchronization, and force interaction realism.

\subsection{Subject-Centric Fine-grained Traits}

\textbf{Subject-Video Quality.} 
We evaluate subject-level video attributes via three metrics: id fidelity, subject attribute consistency, and lip sync. For id fidelity, we employ ArcFace \cite{ArcFace} to extract facial features and compute similarity between generated and reference faces, ensuring alignment with human perception. Subject attribute consistency leverages Gemini-3 to assess the subject's continuous presence and physical coherence over time, focusing on issues such as disappearance, missing or severed body parts, and abrupt facial or body mutations. Lip sync uses SyncNet \cite{syncnet} to measure synchronization accuracy between lip movements and speech, ensuring precise lip-audio alignment. \\
\textbf{Subject-Audio Quality.} We assess subject-level audio attributes via two metrics: pronunciation accuracy and speech realism. The former is measured by computing word error rate (WER) using SenseVoice \cite{SenseVoice} against ground-truth text, quantifying the accuracy and intelligibility of spoken content. speech quality is assessed using the OVRL (Overall Quality) score from DNSMOS \cite{DNSMOS}, a robust non-intrusive metric that evaluates perceptual quality. It effectively captures both the suppression of background noise and the naturalness of the synthesized speech, providing an objective measure aligned with human perception.

\begin{table}[t]
\centering
\caption{\textbf{Evaluation on I2AV task in global level metrics.}}
\vspace{0.05in}
\resizebox{\columnwidth}{!}{%
\renewcommand{\arraystretch}{1.1} 
\setlength{\tabcolsep}{2.8mm} 
\begin{tabular}{l ccc ccc cc}
    \toprule
    \multirow{2}{*}{\textbf{Models}} & \multicolumn{3}{c}{\textbf{Video Quality}} & \multicolumn{3}{c}{\textbf{Audio Quality}} & \multicolumn{2}{c}{\textbf{Multi-modal Align}} \\
    \cmidrule(lr){2-4} \cmidrule(lr){5-7} \cmidrule(lr){8-9}
    
    & IQ$\uparrow$ & DD$\uparrow$ & BP$\uparrow$ & AbS$\uparrow$ & KL$\downarrow$ & FD$\downarrow$ & T-A$\uparrow$ & V-A$\uparrow$ \\
    \midrule
    
    \rowcolor{gray!10} \multicolumn{9}{c}{\textit{\textbf{Closed-source model}}} \\
    \midrule
    \textbf{Veo3} & \underline{\textbf{0.737}} & 0.878 & \underline{\textbf{8.74}} & 5.40 & \underline{\textbf{0.79}} & 0.87 & \underline{0.39} & \underline{\textbf{0.37}} \\
    \textbf{kling2.6} & 0.722 & 0.686 & 8.21 & 4.05 & 1.28 & 1.02 & 0.36 & 0.26\\
    \textbf{Wan2.5} & 0.707 & 0.721 & 8.45 & 4.42 & 1.04 & 0.87 & 0.33 & 0.21 \\
    \textbf{Sora2} & 0.708 & 0.538 & 8.52 & 3.34 & 0.87 & 0.87 & 0.32 & 0.30  \\
    \textbf{SeedDance1.5-pro} & 0.726 & \underline{\textbf{0.897}} & 8.65 & \underline{\textbf{5.63}} & 0.84 & \underline{\textbf{0.83}} & 0.36 & 0.31 \\
    \midrule
    
    \rowcolor{gray!10} \multicolumn{9}{c}{\textit{\textbf{Open-source model}}} \\
    \midrule
    \textbf{Universe-1} & 0.681 & 0.390 & 7.19 & 2.51 & 1.18 & 1.17 & 0.16 & 0.19 \\
    \textbf{Uniavgen} & 0.709 & 0.575 & 8.24 & 4.21 & \underline{0.84} & \underline{0.89} & 0.37 & 0.19 \\
    \textbf{Ovi} & 0.701 & \underline{0.691} & 8.58 & 3.75 & 0.97 & 0.90 & \underline{\textbf{0.40}} & 0.23\\
    \textbf{MOVA} & 0.695 & 0.564 & 8.33 & \underline{4.27} & 0.96 & 1.01 & 0.20 & \underline{0.29} \\
    
    \rowcolor{orange!10} \textbf{LTX-2} & \underline{0.720} & 0.601 & \underline{8.73} & 3.69 & 1.04 & 1.06 & 0.28 & 0.27 \\
    
    \rowcolor{orange!20} & 0.721 & 0.665 & 8.78 & 4.13 & 0.77 & 0.93 & 0.35 & 0.30 \\[-0.2em]
    \rowcolor{orange!20} \multirow{-2}{*}{\textbf{ours(LTX-ft)}} & \scriptsize(+0.1\%) & \scriptsize(+10.7\%) & \scriptsize(+0.5\%) & \scriptsize(+11.9\%) & \scriptsize(+25.9\%) & \scriptsize(+12.3\%) & \scriptsize(+25.0\%) & \scriptsize(+11.1\%) \\
    \bottomrule
\end{tabular}}
\label{tab:global_eval}
\end{table}

\begin{table}[t]
\centering
\caption{\textbf{Evaluation on I2AV task in interaction and individual level metrics.}}
\vspace{0.05in}
\resizebox{\columnwidth}{!}{%
\renewcommand{\arraystretch}{1.1} 
\setlength{\tabcolsep}{1.4mm} 
\begin{tabular}{l cccc cc ccc cc}
    \toprule
    \multirow{2}{*}{\textbf{Models}} & \multicolumn{4}{c}{\textbf{Person-Person}} & \multicolumn{2}{c}{\textbf{Person-Object}} & \multicolumn{3}{c}{\textbf{Subject-Video}} & \multicolumn{2}{c}{\textbf{Subject-Audio}} \\
    \cmidrule(lr){2-5} \cmidrule(lr){6-7} \cmidrule(lr){8-10} \cmidrule(lr){11-12}
    
    & IC$^*$$\uparrow$ & IN$\uparrow$ & LR$\uparrow$ & ES$\uparrow$ & OC$\uparrow$ & CN$\uparrow$ & IC$\uparrow$ & AC$\uparrow$ & Sync$\uparrow$ & SQ$\uparrow$ & WER$\downarrow$ \\
    \midrule
    
    \rowcolor{gray!10} \multicolumn{12}{c}{\textit{\textbf{Closed-source model}}} \\
    \midrule
    \textbf{Veo3} & 0.580 & \underline{\textbf{7.00}} & \underline{8.84} & \underline{7.31} & 7.05 & 8.12 & 0.675 & \underline{\textbf{9.77}} & \underline{\textbf{8.32}} & 2.96 &0.24 \\
    \textbf{kling2.6} & 0.660 & 5.50 & 8.13 & 5.97 & \underline{\textbf{7.76}} & \underline{\textbf{8.32}} & 0.724 & 9.06 & 7.91 & 2.71 &  \underline{\textbf{0.23}} \\
    \textbf{Wan2.5} & 0.705 & 6.02 & 7.97 & 6.52 & 7.24 & 7.86 & \underline{\textbf{0.802}} & 9.45 & 6.83 & 2.59 & 0.24 \\
    \textbf{Sora2} & 0.579 & 6.15 & 7.91 & 6.29 & 7.03 & 7.24 & 0.690 & 9.74 & 7.47 & 2.40 & 0.30\\
    \textbf{SeedDance1.5-pro} & \underline{\textbf{0.714}} & 6.59 & 8.54 & 6.97 & 6.56 & 6.82 & 0.758 & 9.62 & 7.93 & \textbf{\underline{3.02}} & 0.29 \\
    \midrule
    
    \rowcolor{gray!10} \multicolumn{12}{c}{\textit{\textbf{Open-source model}}} \\
    \midrule
    \textbf{Universe-1} & - & - & - & - & 4.72 & 4.97 & \underline{0.683} & 9.41 & 1.75 & 2.56 & 0.76 \\
    \textbf{Uniavgen} & - & - & - & - & 5.24 & 5.36 & 0.599 & 9.43 & 4.45 & \underline{3.01} & 0.30 \\
    \textbf{Ovi}   & 0.443 & 5.84 & 6.35 & \underline{\textbf{7.67}} & 6.61 & 6.72 & 0.588 & \underline{9.61} & \underline{6.97} & 2.96 & 0.29\\
    \textbf{MOVA}     & 0.561 & 6.09 & 7.81 & 6.36 & 5.98 & 5.37 & 0.682 & 9.24 & 6.18 & 2.99 & \underline{0.26} \\
    \rowcolor{orange!10} \textbf{LTX-2}    & \underline{0.562} & \underline{6.47} & \underline{\textbf{8.94}} & 6.63 & \underline{6.95} & \underline{6.98} & 0.656 & 9.51 & 6.91 & 2.89 & 0.29 \\
    
    \rowcolor{orange!20} & 0.589 & 6.53 & 8.92 & 6.76 & 7.12 & 7.32 & 0.696 & 9.64 & 7.11 & 2.98 & 0.27 \\[-0.2em]
    \rowcolor{orange!20} \multirow{-2}{*}{\textbf{ours(LTX-ft)}} & 
    \scriptsize(+4.8\%) & \scriptsize(+0.9\%) & \scriptsize(-0.1\%) & \scriptsize(+2.1\%) & 
    \scriptsize(+2.4\%) & \scriptsize(+4.9\%) & \scriptsize(+6.1\%) & \scriptsize(+1.4\%) & 
    \scriptsize(+2.9\%) & \scriptsize(+3.1\%) & \scriptsize(+6.9\%) \\ 
    \bottomrule
\end{tabular}}
\label{tab:relational_individual_eval}
\end{table}
\begin{table*}[t]
\centering
\caption{\textbf{Evaluation on S2V task: global quality, relational interaction, and individual subject performance.}}
\vspace{0.05in}
\resizebox{\textwidth}{!}{%
\renewcommand{\arraystretch}{1.1} 
\setlength{\tabcolsep}{1.5mm} 
\begin{tabular}{l ccc c | cccc cc | ccc} 
    \toprule
    \multirow{3}{*}{\textbf{Models}} & \multicolumn{4}{c|}{\textbf{Global Level}} & \multicolumn{6}{c|}{\textbf{Relational Level}} & \multicolumn{3}{c}{\textbf{Individual Level}} \\ 
    \cmidrule(lr){2-5} \cmidrule(lr){6-11} \cmidrule(lr){12-14} 
    
    & \multicolumn{3}{c}{\textbf{Video Quality}} & \multicolumn{1}{c|}{\textbf{AV-Align}} & \multicolumn{4}{c}{\textbf{Person-Person}} & \multicolumn{2}{c|}{\textbf{Person-Object}} & \multicolumn{3}{c}{\textbf{Subject Video}} \\ 
    \cmidrule(lr){2-4} \cmidrule(lr){5-5} \cmidrule(lr){6-9} \cmidrule(lr){10-11} \cmidrule(lr){12-14} 
    
    & IQ$\uparrow$ & DD$\uparrow$ & BP$\uparrow$ & V-A$\uparrow$ & IC$^*$$\uparrow$ & IN$\uparrow$ & LR$\uparrow$ & ES$\uparrow$ & OC$\uparrow$ & CN$\uparrow$ & IC$\uparrow$ & AC$\uparrow$ & Sync$\uparrow$ \\ 
    \midrule
    
    \rowcolor{gray!10} \multicolumn{14}{c}{\textit{\textbf{Closed-source model}}} \\ 
    \midrule
    \textbf{kling-avatar\cite{ding2025kling}} & 0.707 & 0.517 & 8.45 & 0.369 & -& -& -& -&\textbf{6.52} & 6.43 & \textbf{0.844}& \textbf{9.81} & 6.89\\
    \textbf{wan2.2-s2v} & \textbf{0.710} & 0.754 & 8.73 & \textbf{0.374} & -& - & - & - &6.41 & 6.32 &0.710 & 9.62 &7.24 \\
    \textbf{Omniavtar} & 0.703 & 0.352 & 8.66 & 0.369 & - & -& -& -& 6.17 & \textbf{6.56} & 0.655 & 9.73 & 6.88 \\
    \textbf{hunyuanvideo-avatar\cite{chen2025hunyuanvideo}} & 0.681 & \textbf{0.830} & 8.40 & 0.351 &- & -& -& -& 5.01 & 5.24 & 0.642 & 9.29 & 5.52 \\
    \textbf{Multitalk} & 0.700 & 0.433 & 8.86 & 0.362 & 0.578 & 5.90 & 7.64 & 6.26 & 6.32 & 6.49 & 0.704 & 9.80 & 7.59 \\
    \textbf{InfiniteTalk} & 0.702 & 0.545 & \textbf{8.87} & 0.369 & \textbf{0.679} & \textbf{6.06} & \textbf{7.94} & \textbf{6.31} & 6.43 & 6.51 & 0.775 & \textbf{9.81} & \textbf{7.63}\\
    \bottomrule
\end{tabular}}
\label{tab:s2v_evaluation}
\end{table*}

\section{Experiment}
\label{sec:experiment}

\subsection{Evaluation details}
We evaluated recent open and closed source models on the I2AV task using the seven-dimensional OHBench. The closed-source models include Veo3.1, Wan2.5, Sora2, kling2.6\cite{kling}, and SeedDance-1.5-pro\cite{seedance1.5pro}; the open-source models include Universe-1\cite{universe}, UniAVGen\cite{uniavgen}, Ovi, LTX-2, and MOVA\cite{ChemCoTBench}. For inference, we generated 5-second videos (or the closest available length for models not supporting 5 seconds). All models were evaluated at the resolution closest to 720P.

\subsection{Main Results}


\textbf{Overall Performance Landscape.} 
We show comprehensive OHBench results for all evaluated models in Tab.~\ref{tab:global_eval} and Tab.~\ref{tab:relational_individual_eval}. Closed-source models generally outperform open-source ones across most metrics. Veo3 and SeedDance1.5-pro lead in overall video quality, interaction rationality, and individual attributes. 

\textbf{Closd-source Model Results.}Veo3 achieves SOTA performance across nearly all metrics, excelling particularly in audio-visual alignment and lip-sync accuracy. SeedDance1.5-pro ranks second overall, with top performance in motion dynamics, audio aesthetics, perceptual similarity, and voice quality. kling2.6 shows a clear advantage in person-object interaction and speech accuracy. Wan2.5 achieves the best identity preservation but lags in speech quality, multi-modal alignment and lip-sync compared to other closed-source models. Sora2 performs relatively poorly among closed-source models, especially in speech quality.

\begin{figure}[t]
    \centering
    \includegraphics[width=0.9\linewidth]{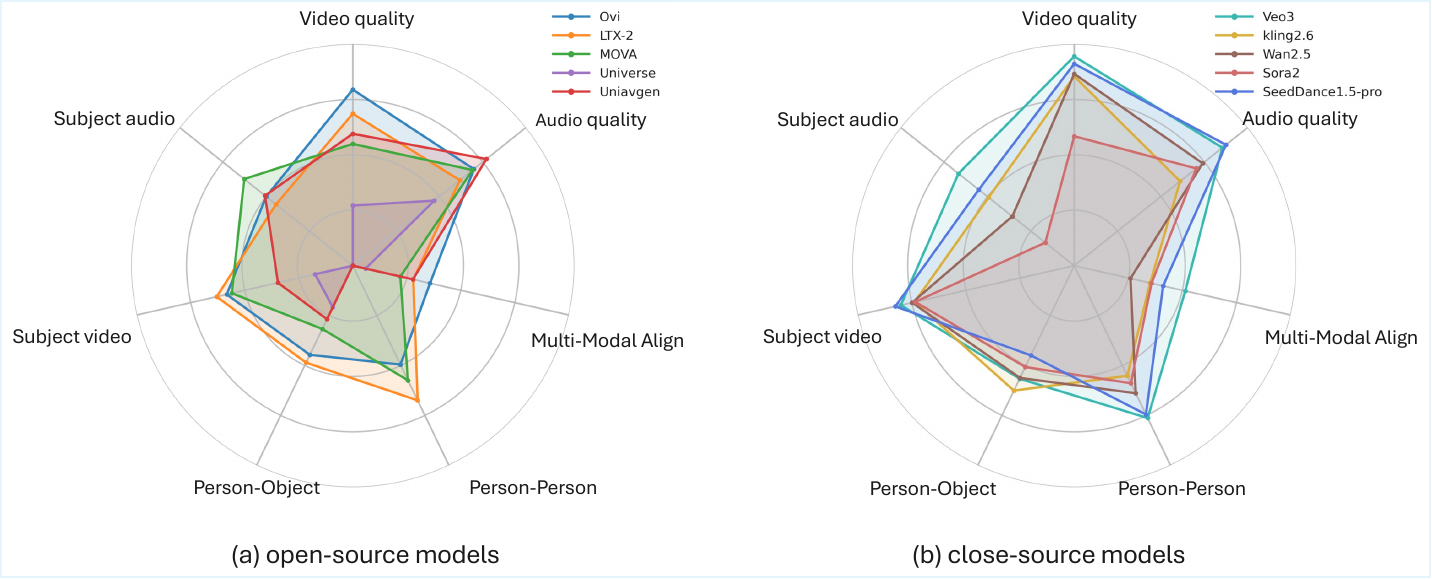}
    \vspace{-1.0em}
    \captionof{figure}{
    Performance distribution of 10 models across seven dimensions on OHBench for audio-video joint generation task.
    }
    \label{fig:rader}
    \vspace{-1.5em}
\end{figure}

\textbf{Open-source Model Results.}
Regarding open-source models, Ovi rivals closed-source models in cross-modal semantic consistency and lip-sync. However, it underperforms in listener realism and occasionally exhibits audio-visual assignment errors (see supplementary material). LTX-2 emerges as the strongest open-source model overall, particularly in dyadic interaction and video quality; notably, its listener realism surpasses all closed-source models. MOVA shows balanced performance across most metrics but struggles with person-object interactions and individual attribute consistency, occasionally producing artifacts like subject disappearance or flickering. UniAVGen achieves competitive results in audio quality and subject attributes but lacks dyadic interaction capability due to insufficient multi-person and person-object training data. Universe-1, as an earlier baseline, lags behind across most evaluation dimensions.\\
\textbf{Capabilities and Limitations Across Dimensions.}
To provide a more intuitive comparison between open-source and closed-source models, we summarize their comprehensive performance across seven dimensions. As illustrated in Fig.~\ref{fig:rader}, existing models perform reasonably in global audio-visual quality, with closed-source models excelling in global video quality. However, individual-level speech and video quality lag behind overall metrics. Moreover, person-person, person-object interaction and multi-modal alignment, remain areas requiring significant improvement.

\textbf{Open-source vs. Closed-source Models.}
Closed-source models significantly outperform open-source ones, particularly in video quality and person-object interaction, while also maintaining a marginal lead in person-person interaction, multi-modal alignment, and audio quality. This superiority stems from their massive training datasets, providing sufficient support across scene-level, interaction-level, and individual-level tasks. Additionally, application-driven optimization prompts these models to incorporate post-training alignment with human preferences. Nevertheless, even for closed-source models, substantial potential for enhancement persists in the quality of multi-modal alignment and person-object interaction.

\textbf{Generalists vs. Specialists: The Impact of Data Strategies and Backbone Designs on Model Quality.}
Notably, Fig.~\ref{fig:rader} reveals that open-source models exhibit more balanced performance, whereas closed-source models show uneven distributions, a phenomenon
closely tied to the architectural trade-offs and data-centric strategies of open-source development.
Specifically, Ovi demonstrates remarkable results in text-audio (T-A) synchronization within multi-modal alignment, even surpassing all closed-source models in the latter. This advantage is attributed to its large-scale audio pre-training. However, Ovi's dual-person identity consistency and listener realism (LR) are limited by scarce complex interaction samples in its training data. Conversely, LTX-2 demonstrates a vital stunning capability in the person to person
interaction, especially in listner realism, slightly outperforming Veo3. It also 
maintains a significant lead in overall video quality (IQ). However, its audio quality suffers due 
\begin{wrapfigure}[18]{r}{0.42\columnwidth}
    \vspace{-2mm}
    \centering
    \includegraphics[width=\linewidth]{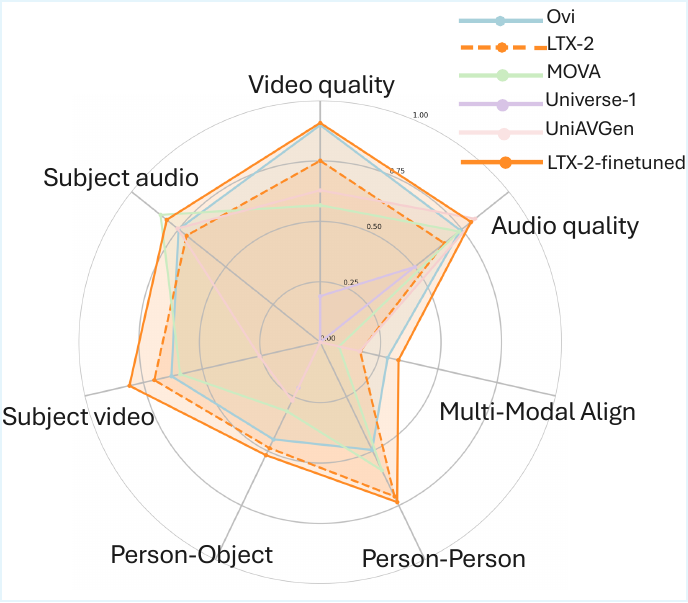}
    \caption{\footnotesize Performance comparison of LTX-2 before and after finetuning on the OmniHuman data subset.}
    \label{fig:ft_rader}
    \vspace{-3mm}
\end{wrapfigure}
to its audio branch's 5B parameters and highly compression Mel-spectrogram latent space, prioritizing speed over acoustic fidelity; training data quality is also be a limiting factor.

\textbf{Our more Observations.} We observe that model performance on human generation significantly degrades as the shot scale transitions from close-up to long shot. Both lip movements during speech and overall body motions become
worse, with increased artifacts in the mouth region and facial distortions. This is likely attributable to the inherent difficulty of modeling humans in distant views and the imbalanced distribution of shot types. Additionally, we find that identity distortion rates are higher in two-person scenarios. Detailed case analyses are provided in the supplementary material. \\
\textbf{More Application.}
For the task of speech-driven video generation, we selected appropriate evaluation metrics from OHBench and systematically assessed current state-of-the-art methods. The results are presented in Tab.~\ref{tab:s2v_evaluation}. Overall, InfiniteTalk achieves the best performance across most metrics. Due to space limitations, additional results and detailed analysis are provided in the supplementary material.

\subsection{Fine-tuning with OmniHuman improves quality.}


We randomly sampled 180,000 single-person and 20,000 two-person samples from OmniHuman dataset to finetune LTX-2, a representative open-source model. As shown in the last rows of Tab.~\ref{tab:global_eval} and Tab.~\ref{tab:relational_individual_eval}, the finetuned model achieves improvements across all evaluation metrics. Notably, audio quality shows significant gains (KL+25.9\%, FD+12.3\%, AbS+11.9\%), and multi-modal alignment is also substantially enhanced (T-A+25.0\%, V-A+11.1\%). Furthermore, clear improvements are observed in dynamic degree (DD+10.7\%), identity consistency (IC+6.1\%, IC$^*$+4.8\%), and contact naturalness (CN+4.9\%).

Fig.~\ref{fig:ft_rader} illustrates the performance gains from fine-tuning across seven dimensions. The results demonstrate that fine-tuning effectively compensates for LTX-2's original weaknesses in audio quality, multi-modal alignment, and subject-audio attributes, while further amplifying its advantages in subject Video, person object, and person-person dimensions. These results demonstrate that the high-quality audio-visual pairs from OmniHuman allows LTX-2 to transcend its original generation upper bound, establishing it as a leading open-source model.
\section{Conclusion}
\label{sec:conclusion}
We presented OmniHuman and OHBench to advance human-centric video-audio joint generation. By addressing structural data deficiencies, we introduced a hierarchical framework spanning global, relational, and individual realism. Our work offers a large-scale dataset with video/frame/individual-level annotations via an automated pipeline. Complementarily, OHBench establishes perception-consistent evaluation by filtering out misaligned metrics and incorporating missing dimensions. Experiments confirm OmniHuman boosts performance with minimal data, while our benchmark aligns with human judgment.

\clearpage  

%
%
\bibliographystyle{splncs04}
\bibliography{main}

@String(ICASSP=	{ICASSP})

@misc{kling,
  title={Kling}, 
  author={Kling},
  year={2025},
  howpublished={\url{https://klingai.com}}
}

@misc{veo3.1,
  author = {{Google DeepMind}},
  title  = {Veo 3},
  year   = {2025},
  month  = {5},
  url    = {https://deepmind.google/models/veo/},
  note   = {Accessed: 2026-02-17}
}

@misc{openai2025sora2,
  author = {OpenAI},
  title  = {Sora 2: Video Generation Model},
  year   = {2025},
  url = {https://openai.com/sora},
}

@article{wang2024av,
  title={Av-dit: Efficient audio-visual diffusion transformer for joint audio and video generation},
  author={Wang, Kai and Deng, Shijian and Shi, Jing and Hatzinakos, Dimitrios and Tian, Yapeng},
  journal={arXiv preprint arXiv:2406.07686},
  year={2024}
}

@article{yang2025infinitetalk,
  title={Infinitetalk: Audio-driven video generation for sparse-frame video dubbing},
  author={Yang, Shaoshu and Kong, Zhe and Gao, Feng and Cheng, Meng and Liu, Xiangyu and Zhang, Yong and Kang, Zhuoliang and Luo, Wenhan and Cai, Xunliang and He, Ran and others},
  journal={arXiv preprint arXiv:2508.14033},
  year={2025}
}

@article{yang2025unimatch,
  title={Unimatch v2: Pushing the limit of semi-supervised semantic segmentation},
  author={Yang, Lihe and Zhao, Zhen and Zhao, Hengshuang},
  journal={IEEE Transactions on Pattern Analysis and Machine Intelligence},
  volume={47},
  number={4},
  pages={3031--3048},
  year={2025},
  publisher={IEEE}
}

@article{sigclip,
  title={Siglip 2: Multilingual vision-language encoders with improved semantic understanding, localization, and dense features},
  author={Tschannen, Michael and Gritsenko, Alexey and Wang, Xiao and Naeem, Muhammad Ferjad and Alabdulmohsin, Ibrahim and Parthasarathy, Nikhil and Evans, Talfan and Beyer, Lucas and Xia, Ye and Mustafa, Basil and others},
  journal={arXiv preprint arXiv:2502.14786},
  year={2025}
}

@inproceedings{iashin2024synchformer,
  title={Synchformer: Efficient synchronization from sparse cues},
  author={Iashin, Vladimir and Xie, Weidi and Rahtu, Esa and Zisserman, Andrew},
  booktitle={ICASSP 2024-2024 IEEE International Conference on Acoustics, Speech and Signal Processing (ICASSP)},
  pages={5325--5329},
  year={2024},
  organization={IEEE}
}

@inproceedings{clip,
  title={Learning transferable visual models from natural language supervision},
  author={Radford, Alec and Kim, Jong Wook and Hallacy, Chris and Ramesh, Aditya and Goh, Gabriel and Agarwal, Sandhini and Sastry, Girish and Askell, Amanda and Mishkin, Pamela and Clark, Jack and others},
  booktitle={International conference on machine learning},
  pages={8748--8763},
  year={2021},
  organization={PmLR}
}

@article{di2023magicdance,
  title={Magicdance: Realistic human dance video generation with motions \& facial expressions transfer},
  author={Di Chang, Yichun Shi and Gao, Quankai and Fu, Jessica and Xu, Hongyi and Song, Guoxian and Yan, Qing and Yang, Xiao and Soleymani, Mohammad},
  journal={arXiv preprint arXiv:2311.12052},
  volume={2},
  number={3},
  pages={4},
  year={2023}
}

@inproceedings{caba2015activitynet,
  title={Activitynet: A large-scale video benchmark for human activity understanding},
  author={Caba Heilbron, Fabian and Escorcia, Victor and Ghanem, Bernard and Carlos Niebles, Juan},
  booktitle={Proceedings of the ieee conference on computer vision and pattern recognition},
  pages={961--970},
  year={2015}
}

@article{ding2025kling,
  title={Kling-avatar: Grounding multimodal instructions for cascaded long-duration avatar animation synthesis},
  author={Ding, Yikang and Liu, Jiwen and Zhang, Wenyuan and Wang, Zekun and Hu, Wentao and Cui, Liyuan and Lao, Mingming and Shao, Yingchao and Liu, Hui and Li, Xiaohan and others},
  journal={arXiv preprint arXiv:2509.09595},
  year={2025}
}

@article{chen2025hunyuanvideo,
  title={Hunyuanvideo-avatar: High-fidelity audio-driven human animation for multiple characters},
  author={Chen, Yi and Liang, Sen and Zhou, Zixiang and Huang, Ziyao and Ma, Yifeng and Tang, Junshu and Lin, Qin and Zhou, Yuan and Lu, Qinglin},
  journal={arXiv preprint arXiv:2505.20156},
  year={2025}
}

@article{gan2025omniavatar,
  title={Omniavatar: Efficient audio-driven avatar video generation with adaptive body animation},
  author={Gan, Qijun and Yang, Ruizi and Zhu, Jianke and Xue, Shaofei and Hoi, Steven},
  journal={arXiv preprint arXiv:2506.18866},
  year={2025}
}

@article{sung2024multitalk,
  title={Multitalk: Enhancing 3d talking head generation across languages with multilingual video dataset},
  author={Sung-Bin, Kim and Chae-Yeon, Lee and Son, Gihun and Hyun-Bin, Oh and Ju, Janghoon and Nam, Suekyeong and Oh, Tae-Hyun},
  journal={arXiv preprint arXiv:2406.14272},
  year={2024}
}

@article{uniavgen,
  title={Uniavgen: Unified audio and video generation with asymmetric cross-modal interactions},
  author={Zhang, Guozhen and Zhou, Zixiang and Hu, Teng and Peng, Ziqiao and Zhang, Youliang and Chen, Yi and Zhou, Yuan and Lu, Qinglin and Wang, Limin},
  journal={arXiv preprint arXiv:2511.03334},
  year={2025}
}

@article{apollo,
  title={Apollo: Unified Multi-Task Audio-Video Joint Generation},
  author={Wang, Jun and Qiang, Chunyu and Guo, Yuxin and Wang, Yiran and Zeng, Xijuan and Deng, Feng},
  journal={arXiv e-prints},
  pages={arXiv--2601},
  year={2026}
}

@article{jova,
  title={JoVA: Unified Multimodal Learning for Joint Video-Audio Generation},
  author={Huang, Xiaohu and Zhou, Hao and Yang, Qiangpeng and Wen, Shilei and Han, Kai},
  journal={arXiv preprint arXiv:2512.13677},
  year={2025}
}

@misc{ovi,
      title={Ovi: Twin Backbone Cross-Modal Fusion for Audio-Video Generation}, 
      author={Chetwin Low and Weimin Wang and Calder Katyal},
      year={2025},
      eprint={2510.01284},
      archivePrefix={arXiv},
      primaryClass={cs.MM},
      url={https://arxiv.org/abs/2510.01284}, 
}

@inproceedings{MOVA,
  title={MoVa: Towards generalizable classification of human morals and values},
  author={Chen, Ziyu and Sun, Junfei and Li, Chenxi and Nguyen, Tuan Dung and Yao, Jing and Yi, Xiaoyuan and Xie, Xing and Tan, Chenhao and Xie, Lexing},
  booktitle={Proceedings of the 2025 Conference on Empirical Methods in Natural Language Processing},
  pages={33204--33248},
  year={2025}
}

@inproceedings{MM-LDM,
  title={Mm-ldm: Multi-modal latent diffusion model for sounding video generation},
  author={Sun, Mingzhen and Wang, Weining and Qiao, Yanyuan and Sun, Jiahui and Qin, Zihan and Guo, Longteng and Zhu, Xinxin and Liu, Jing},
  booktitle={Proceedings of the 32nd ACM International Conference on Multimedia},
  pages={10853--10861},
  year={2024}
}

@misc{LTX-2,
      title={LTX-2: Efficient Joint Audio-Visual Foundation Model}, 
      author={Yoav HaCohen and Benny Brazowski and Nisan Chiprut and Yaki Bitterman and Andrew Kvochko and Avishai Berkowitz and Daniel Shalem and Daphna Lifschitz and Dudu Moshe and Eitan Porat and Eitan Richardson and Guy Shiran and Itay Chachy and Jonathan Chetboun and Michael Finkelson and Michael Kupchick and Nir Zabari and Nitzan Guetta and Noa Kotler and Ofir Bibi and Ori Gordon and Poriya Panet and Roi Benita and Shahar Armon and Victor Kulikov and Yaron Inger and Yonatan Shiftan and Zeev Melumian and Zeev Farbman},
      year={2026},
      eprint={2601.03233},
      archivePrefix={arXiv},
      primaryClass={cs.CV},
      url={https://arxiv.org/abs/2601.03233}, 
}

@article{intergen,
  title={Intergen: Diffusion-based multi-human motion generation under complex interactions},
  author={Liang, Han and Zhang, Wenqian and Li, Wenxuan and Yu, Jingyi and Xu, Lan},
  journal={International Journal of Computer Vision},
  volume={132},
  number={9},
  pages={3463--3483},
  year={2024},
  publisher={Springer}
}

@article{voxceleb2,
  title={Voxceleb2: Deep speaker recognition},
  author={Chung, Joon Son and Nagrani, Arsha and Zisserman, Andrew},
  journal={arXiv preprint arXiv:1806.05622},
  year={2018}
}

@inproceedings{HDTF,
  title={Flow-guided one-shot talking face generation with a high-resolution audio-visual dataset},
  author={Zhang, Zhimeng and Li, Lincheng and Ding, Yu and Fan, Changjie},
  booktitle={Proceedings of the IEEE/CVF conference on computer vision and pattern recognition},
  pages={3661--3670},
  year={2021}
}

@inproceedings{li2025openhumanvid,
  title={Openhumanvid: A large-scale high-quality dataset for enhancing human-centric video generation},
  author={Li, Hui and Xu, Mingwang and Zhan, Yun and Mu, Shan and Li, Jiaye and Cheng, Kaihui and Chen, Yuxuan and Chen, Tan and Ye, Mao and Wang, Jingdong and others},
  booktitle={Proceedings of the Computer Vision and Pattern Recognition Conference},
  pages={7752--7762},
  year={2025}
}

@article{yang2025svbench,
  title={Svbench: A benchmark with temporal multi-turn dialogues for streaming video understanding},
  author={Yang, Zhenyu and Hu, Yuhang and Du, Zemin and Xue, Dizhan and Qian, Shengsheng and Wu, Jiahong and Yang, Fan and Dong, Weiming and Xu, Changsheng},
  journal={arXiv preprint arXiv:2502.10810},
  year={2025}
}

@article{zhou2026mtavg,
  title={MTAVG-Bench: A Comprehensive Benchmark for Evaluating Multi-Talker Dialogue-Centric Audio-Video Generation},
  author={Zhou, Yang-Hao and Li, Haitian and Lin, Rexar and Huang, Heyan and Zhou, Jinxing and Yuan, Changsen and Lan, Tian and Zhou, Ziqin and Li, Yudong and Xu, Jiajun and others},
  journal={arXiv preprint arXiv:2602.00607},
  year={2026}
}

@article{xie2025phyavbench,
  title={PhyAVBench: A Challenging Audio Physics-Sensitivity Benchmark for Physically Grounded Text-to-Audio-Video Generation},
  author={Xie, Tianxin and Lei, Wentao and Huang, Guanjie and Zhang, Pengfei and Jiang, Kai and Zhang, Chunhui and Ma, Fengji and He, Haoyu and Zhang, Han and He, Jiangshan and others},
  journal={arXiv preprint arXiv:2512.23994},
  year={2025}
}

@article{hua2025vabench,
  title={Vabench: A comprehensive benchmark for audio-video generation},
  author={Hua, Daili and Wang, Xizhi and Zeng, Bohan and Huang, Xinyi and Liang, Hao and Niu, Junbo and Chen, Xinlong and Xu, Quanqing and Zhang, Wentao},
  journal={arXiv preprint arXiv:2512.09299},
  year={2025}
}

@article{speakervid,
  title={Speakervid-5m: A large-scale high-quality dataset for audio-visual dyadic interactive human generation},
  author={Zhang, Youliang and Li, Zhaoyang and Wang, Duomin and Zhang, Jiahe and Zhou, Deyu and Yin, Zixin and Dai, Xili and Yu, Gang and Li, Xiu},
  journal={arXiv preprint arXiv:2507.09862},
  year={2025}
}

@inproceedings{celebV,
  title={CelebV-HQ: A large-scale video facial attributes dataset},
  author={Zhu, Hao and Wu, Wayne and Zhu, Wentao and Jiang, Liming and Tang, Siwei and Zhang, Li and Liu, Ziwei and Loy, Chen Change},
  booktitle={European conference on computer vision},
  pages={650--667},
  year={2022},
  organization={Springer}
}

@inproceedings{vbench,
  title={Vbench: Comprehensive benchmark suite for video generative models},
  author={Huang, Ziqi and He, Yinan and Yu, Jiashuo and Zhang, Fan and Si, Chenyang and Jiang, Yuming and Zhang, Yuanhan and Wu, Tianxing and Jin, Qingyang and Chanpaisit, Nattapol and others},
  booktitle={Proceedings of the IEEE/CVF Conference on Computer Vision and Pattern Recognition},
  pages={21807--21818},
  year={2024}
}

@article{vabench,
  title={Vabench: A comprehensive benchmark for audio-video generation},
  author={Hua, Daili and Wang, Xizhi and Zeng, Bohan and Huang, Xinyi and Liang, Hao and Niu, Junbo and Chen, Xinlong and Xu, Quanqing and Zhang, Wentao},
  journal={arXiv preprint arXiv:2512.09299},
  year={2025}
}

@article{audiobox,
  title={Meta audiobox aesthetics: Unified automatic quality assessment for speech, music, and sound},
  author={Tjandra, Andros and Wu, Yi-Chiao and Guo, Baishan and Hoffman, John and Ellis, Brian and Vyas, Apoorv and Shi, Bowen and Chen, Sanyuan and Le, Matt and Zacharov, Nick and others},
  journal={arXiv preprint arXiv:2502.05139},
  year={2025}
}

@inproceedings{MUSIQ,
  title={Musiq: Multi-scale image quality transformer},
  author={Ke, Junjie and Wang, Qifei and Wang, Yilin and Milanfar, Peyman and Yang, Feng},
  booktitle={Proceedings of the IEEE/CVF international conference on computer vision},
  pages={5148--5157},
  year={2021}
}

@inproceedings{RAFT,
  title={Raft: Recurrent all-pairs field transforms for optical flow},
  author={Teed, Zachary and Deng, Jia},
  booktitle={European conference on computer vision},
  pages={402--419},
  year={2020},
  organization={Springer}
}

@article{universe,
  title={UniVerse-1: Unified Audio-Video Generation via Stitching of Experts},
  author={Wang, Duomin and Zuo, Wei and Li, Aojie and Chen, Ling-Hao and Liao, Xinyao and Zhou, Deyu and Yin, Zixin and Dai, Xili and Jiang, Daxin and Yu, Gang},
  journal={arXiv preprint arXiv:2509.06155},
  year={2025}
}

@article{javisdit,
  title={Javisdit: Joint audio-video diffusion transformer with hierarchical spatio-temporal prior synchronization},
  author={Liu, Kai and Li, Wei and Chen, Lai and Wu, Shengqiong and Zheng, Yanhao and Ji, Jiayi and Zhou, Fan and Jiang, Rongxin and Luo, Jiebo and Fei, Hao and others},
  journal={arXiv preprint arXiv:2503.23377},
  year={2025}
}

@inproceedings{CLAP,
  title={Large-scale contrastive language-audio pretraining with feature fusion and keyword-to-caption augmentation},
  author={Wu, Yusong and Chen, Ke and Zhang, Tianyu and Hui, Yuchen and Berg-Kirkpatrick, Taylor and Dubnov, Shlomo},
  booktitle={ICASSP 2023-2023 IEEE International Conference on Acoustics, Speech and Signal Processing (ICASSP)},
  pages={1--5},
  year={2023},
  organization={IEEE}
}

@inproceedings{ImageBind,
  title={Large-scale contrastive language-audio pretraining with feature fusion and keyword-to-caption augmentation},
  author={Wu, Yusong and Chen, Ke and Zhang, Tianyu and Hui, Yuchen and Berg-Kirkpatrick, Taylor and Dubnov, Shlomo},
  booktitle={ICASSP 2023-2023 IEEE International Conference on Acoustics, Speech and Signal Processing (ICASSP)},
  pages={1--5},
  year={2023},
  organization={IEEE}
}

@inproceedings{Transnetv2,
  title={Transnet v2: An effective deep network architecture for fast shot transition detection},
  author={Soucek, Tom{\'a}s and Lokoc, Jakub},
  booktitle={Proceedings of the 32nd ACM International Conference on Multimedia},
  pages={11218--11221},
  year={2024}
}

@article{Demucs,
  title={Demucs: Deep extractor for music sources with extra unlabeled data remixed},
  author={D{\'e}fossez, Alexandre and Usunier, Nicolas and Bottou, L{\'e}on and Bach, Francis},
  journal={arXiv preprint arXiv:1909.01174},
  year={2019}
}

@article{DOVER,
  title={Disentangling aesthetic and technical effects for video quality assessment of user generated content},
  author={Wu, Haoning and Liao, Liang and Chen, Chaofeng and Hou, Jingwen and Wang, Annan and Sun, Wenxiu and Yan, Qiong and Lin, Weisi},
  journal={arXiv preprint arXiv:2211.04894},
  volume={2},
  number={5},
  pages={6},
  year={2022}
}

@inproceedings{ArcFace,
  title={Arcface: Additive angular margin loss for deep face recognition},
  author={Deng, Jiankang and Guo, Jia and Xue, Niannan and Zafeiriou, Stefanos},
  booktitle={Proceedings of the IEEE/CVF conference on computer vision and pattern recognition},
  pages={4690--4699},
  year={2019}
}

@article{SenseVoice,
  title={Funaudiollm: Voice understanding and generation foundation models for natural interaction between humans and llms},
  author={An, Keyu and Chen, Qian and Deng, Chong and Du, Zhihao and Gao, Changfeng and Gao, Zhifu and Gu, Yue and He, Ting and Hu, Hangrui and Hu, Kai and others},
  journal={arXiv preprint arXiv:2407.04051},
  year={2024}
}

@article{WeaklyTVQA,
  title={Weakly-supervised 3d spatial reasoning for text-based visual question answering},
  author={Li, Hao and Huang, Jinfa and Jin, Peng and Song, Guoli and Wu, Qi and Chen, Jie},
  journal={IEEE Transactions on Image Processing},
  volume={32},
  pages={3367--3382},
  year={2023},
  publisher={IEEE}
}

@article{YOLOv11,
  title={YOLOv8 to YOLO11: A comprehensive architecture in-depth comparative review},
  author={Hidayatullah, Priyanto and Syakrani, Nurjannah and Sholahuddin, Muhammad Rizqi and Gelar, Trisna and Tubagus, Refdinal},
  journal={arXiv preprint arXiv:2501.13400},
  year={2025}
}

@inproceedings{MOTRv2,
  title={Motrv2: Bootstrapping end-to-end multi-object tracking by pretrained object detectors},
  author={Zhang, Yuang and Wang, Tiancai and Zhang, Xiangyu},
  booktitle={Proceedings of the IEEE/CVF conference on computer vision and pattern recognition},
  pages={22056--22065},
  year={2023}
}

@inproceedings{3Dspeaker,
  title={3D-Speaker-Toolkit: An Open-Source Toolkit for Multimodal Speaker Verification and Diarization},
  author={Chen, Yafeng and Zheng, Siqi and Wang, Hui and Cheng, Luyao and Zhu, Tinglong and Huang, Rongjie and Deng, Chong and Chen, Qian and Zhang, Shiliang and Wang, Wen and others},
  booktitle={ICASSP 2025-2025 IEEE International Conference on Acoustics, Speech and Signal Processing (ICASSP)},
  pages={1--5},
  year={2025},
  organization={IEEE}
}

@inproceedings{syncnet,
  title={Out of time: automated lip sync in the wild},
  author={Chung, Joon Son and Zisserman, Andrew},
  booktitle={Asian conference on computer vision},
  pages={251--263},
  year={2016},
  organization={Springer}
}

@misc{qwen-omni,
      title={Qwen3-Omni Technical Report}, 
      author={Jin Xu and Zhifang Guo and Hangrui Hu and Yunfei Chu and Xiong Wang and Jinzheng He and Yuxuan Wang and Xian Shi and Ting He and Xinfa Zhu and Yuanjun Lv and Yongqi Wang and Dake Guo and He Wang and Linhan Ma and Pei Zhang and Xinyu Zhang and Hongkun Hao and Zishan Guo and Baosong Yang and Bin Zhang and Ziyang Ma and Xipin Wei and Shuai Bai and Keqin Chen and Xuejing Liu and Peng Wang and Mingkun Yang and Dayiheng Liu and Xingzhang Ren and Bo Zheng and Rui Men and Fan Zhou and Bowen Yu and Jianxin Yang and Le Yu and Jingren Zhou and Junyang Lin},
      year={2025},
      eprint={2509.17765},
      archivePrefix={arXiv},
      primaryClass={cs.CL},
      url={https://arxiv.org/abs/2509.17765}, 
}

@inproceedings{DNSMOS,
  title={DNSMOS: A non-intrusive perceptual objective speech quality metric to evaluate noise suppressors},
  author={Reddy, Chandan KA and Gopal, Vishak and Cutler, Ross},
  booktitle={ICASSP 2021-2021 IEEE International Conference on Acoustics, Speech and Signal Processing (ICASSP)},
  pages={6493--6497},
  year={2021},
  organization={IEEE}
}

@article{seedance1.5pro,
  title={Seedance 1.5 pro: A Native Audio-Visual Joint Generation Foundation Model},
  author={Seedance, Team and Chen, Heyi and Chen, Siyan and Chen, Xin and Chen, Yanfei and Chen, Ying and Chen, Zhuo and Cheng, Feng and Cheng, Tianheng and Cheng, Xinqi and others},
  journal={arXiv preprint arXiv:2512.13507},
  year={2025}
}

@article{wan2.5,
  title={Wan: Open and advanced large-scale video generative models},
  author={Wan, Team and Wang, Ang and Ai, Baole and Wen, Bin and Mao, Chaojie and Xie, Chen-Wei and Chen, Di and Yu, Feiwu and Zhao, Haiming and Yang, Jianxiao and others},
  journal={arXiv preprint arXiv:2503.20314},
  year={2025}
}

@inproceedings{FreestyleRet,
  title={Freestyleret: retrieving images from style-diversified queries},
  author={Li, Hao and Jia, Yanhao and Jin, Peng and Cheng, Zesen and Li, Kehan and Sui, Jialu and Liu, Chang and Yuan, Li},
  booktitle={European Conference on Computer Vision},
  pages={258--274},
  year={2024},
  organization={Springer}
}

@article{ChemCoTBench,
  title={Beyond Chemical QA: Evaluating LLM's Chemical Reasoning with Modular Chemical Operations},
  author={Li, Hao and Cao, He and Feng, Bin and Shao, Yanjun and Tang, Xiangru and Yan, Zhiyuan and Yuan, Li and Tian, Yonghong and Li, Yu},
  journal={arXiv preprint arXiv:2505.21318},
  year={2025}
}

@misc{an2025funasrtechnicalreport,
      title={Fun-ASR Technical Report},
      author={Keyu An and Yanni Chen and Zhigao Chen and Chong Deng and Zhihao Du and Changfeng Gao and Zhifu Gao and Bo Gong and Xiangang Li and Yabin Li and Ying Liu and Xiang Lv and Yunjie Ji and Yiheng Jiang and Bin Ma and Haoneng Luo and Chongjia Ni and Zexu Pan and Yiping Peng and Zhendong Peng and Peiyao Wang and Hao Wang and Haoxu Wang and Wen Wang and Wupeng Wang and Yuzhong Wu and Biao Tian and Zhentao Tan and Nan Yang and Bin Yuan and Jieping Ye and Jixing Yu and Qinglin Zhang and Kun Zou and Han Zhao and Shengkui Zhao and Jingren Zhou and Yanqiao Zhu},
      year={2025},
      eprint={2509.12508},
      archivePrefix={arXiv},
      primaryClass={cs.CL},
      url={https://arxiv.org/abs/2509.12508},
}

@inproceedings{peng2024synctalk,
  title={Synctalk: The devil is in the synchronization for talking head synthesis},
  author={Peng, Ziqiao and Hu, Wentao and Shi, Yue and Zhu, Xiangyu and Zhang, Xiaomei and Zhao, Hao and He, Jun and Liu, Hongyan and Fan, Zhaoxin},
  booktitle={Proceedings of the IEEE/CVF Conference on Computer Vision and Pattern Recognition},
  pages={666--676},
  year={2024}
}

@inproceedings{peng2023selftalk,
  title={Selftalk: A self-supervised commutative training diagram to comprehend 3d talking faces},
  author={Peng, Ziqiao and Luo, Yihao and Shi, Yue and Xu, Hao and Zhu, Xiangyu and Liu, Hongyan and He, Jun and Fan, Zhaoxin},
  booktitle={Proceedings of the 31st ACM International Conference on Multimedia},
  pages={5292--5301},
  year={2023}
}

@article{peng2025omnisync,
  title={Omnisync: Towards universal lip synchronization via diffusion transformers},
  author={Peng, Ziqiao and Liu, Jiwen and Zhang, Haoxian and Liu, Xiaoqiang and Tang, Songlin and Wan, Pengfei and Zhang, Di and Liu, Hongyan and He, Jun},
  journal={arXiv preprint arXiv:2505.21448},
  year={2025}
}

@inproceedings{peng2025dualtalk,
  title={Dualtalk: Dual-speaker interaction for 3d talking head conversations},
  author={Peng, Ziqiao and Fan, Yanbo and Wu, Haoyu and Wang, Xuan and Liu, Hongyan and He, Jun and Fan, Zhaoxin},
  booktitle={Proceedings of the Computer Vision and Pattern Recognition Conference},
  pages={21055--21064},
  year={2025}
}
\end{document}